\documentclass[letterpaper, 10 pt, conference]{ieeeconf}  

\IEEEoverridecommandlockouts                              

\title{\LARGE \bf
CooperDrive: Enhancing Driving Decisions Through Cooperative Perception}

\author{Deyuan Qu, Qi Chen, Takayuki Shimizu, Onur Altintas\\
\thanks{The authors are with Toyota InfoTech Labs, Mountain View, CA, USA}%
\thanks{Email:\texttt{\{deyuan.qu, qi.chen, takayuki.shimizu, onur.altintas\}@toyota.com}}%
}

\usepackage{hyperref}
\usepackage{graphicx}
\usepackage{amsmath}
\usepackage{amssymb}
\newcommand{\ignore}[1]{}
\usepackage{arydshln}
\usepackage{colortbl}
\usepackage{xcolor}
\usepackage{booktabs}
\usepackage{tabularx}
\usepackage{makecell}
\usepackage{multirow} 
\usepackage{multicol}
\usepackage{subcaption}
\usepackage{epstopdf}
\usepackage{algorithm}
\usepackage{algpseudocode}
\definecolor{lightpink}{RGB}{255,210,220}
\definecolor{lightgreen}{RGB}{200,255,200}
\definecolor{lightblue}{RGB}{173,216,230}
\begin{document}

\maketitle
\thispagestyle{empty}
\pagestyle{empty}

\begin{abstract}
Autonomous vehicles equipped with robust onboard perception, localization, and planning still face limitations in occlusion and non-line-of-sight (NLOS) scenarios, where delayed reactions can increase collision risk. We propose CooperDrive, a cooperative perception framework that augments situational awareness and enables earlier, safer driving decisions. CooperDrive offers two key advantages: (i) each vehicle retains its native perception, localization, and planning stack, and (ii) a lightweight object-level sharing and fusion strategy bridges perception and planning. Specifically, CooperDrive reuses detector Bird’s-Eye View (BEV) features to estimate accurate vehicle poses without additional heavy encoders, thereby reconstructing BEV representations and feeding the planner with low latency. On the planning side, CooperDrive leverages the expanded object set to anticipate potential conflicts earlier and adjust speed and trajectory proactively, thereby transforming reactive behaviors into predictive and safer driving decisions. Real-world closed-loop tests at occlusion-heavy NLOS intersections demonstrate that CooperDrive increases reaction lead time, minimum time-to-collision (TTC), and stopping margin, while requiring only ~90 kbps bandwidth and maintaining an average end-to-end latency of 89 ms.
\smallskip
\noindent\textbf{Project Page:} \url{https://DarrenQu.github.io/CooperDrive/}
\end{abstract}

\section{Introduction}
\begin{figure}[t]
    \centering
    \includegraphics[width=0.8\linewidth]{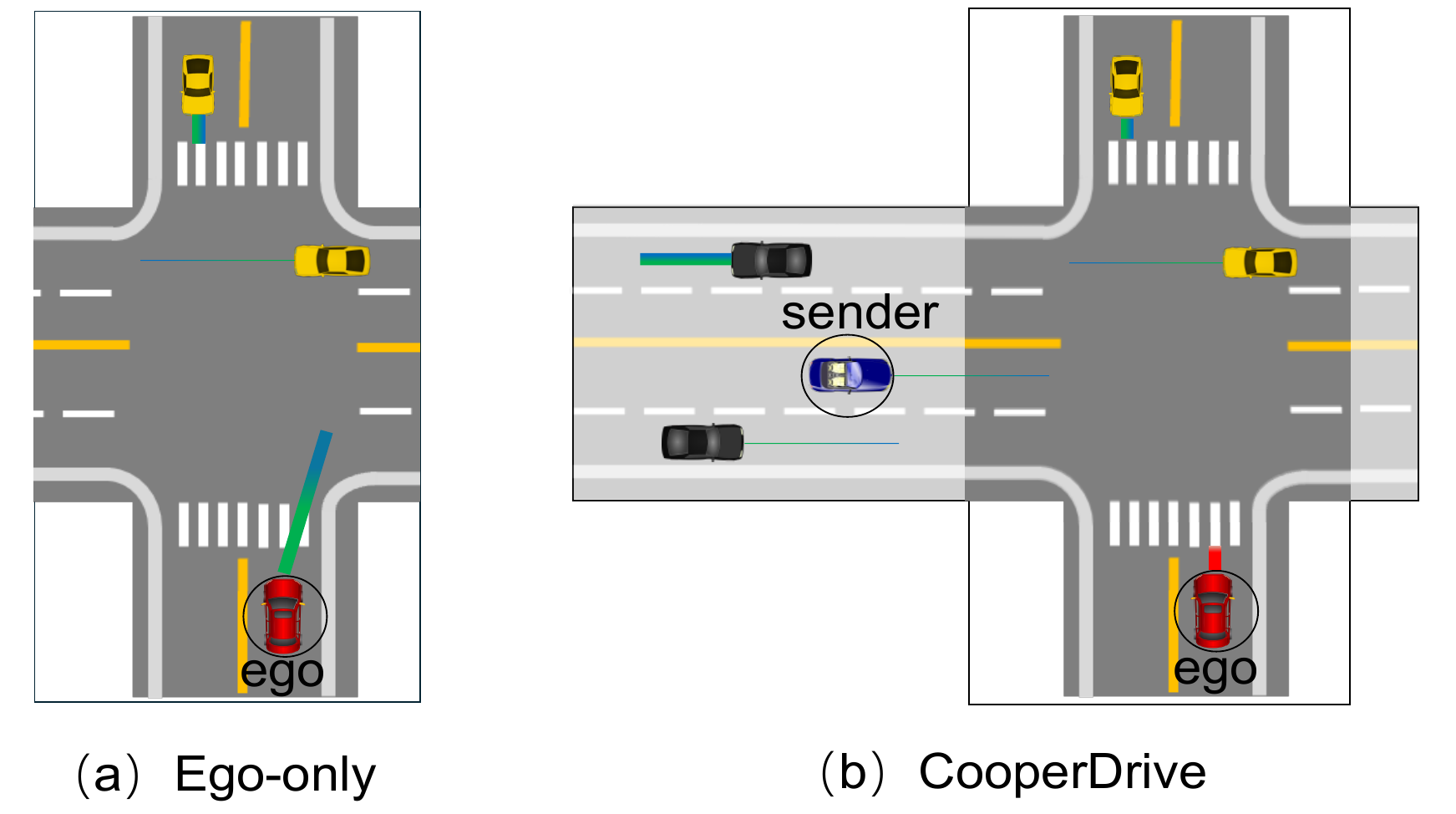}
    \caption{Autonomous vehicles inevitably face sensing limitations due to occlusions and blind spots. Cooperative perception addresses this challenge by allowing vehicles to share complementary observations, enabling earlier awareness of potential hazards and supporting safer driving decisions.}
    \label{fig:1}
    \vspace{-6mm}
\end{figure}

Autonomous driving systems have achieved remarkable progress in recent years, yet they remain fundamentally constrained by the limited field of view of onboard sensors.  Even with high-resolution LiDAR, radar, and multi-camera arrays, an ego vehicle alone cannot fully resolve occlusions or long-range uncertainties, often resulting in delayed hazard recognition and reactive maneuvers in complex urban environments, shown in Figure~\ref{fig:1} (a).

Cooperative perception has therefore emerged as a compelling paradigm: by exchanging information among connected vehicles or infrastructure, it extends the effective perception horizon, mitigates occlusions, and provides a more comprehensive situational awareness in Figure~\ref{fig:1} (b).

Despite this promise, existing cooperative perception studies~\cite{chen2019cooper, chen2019f, xu2022v2x, xu2022cobevt, qu2024sicp} remain limited in scope. Most prior work focuses on perception-only benchmarks, such as improving detection or tracking accuracy via early fusion, deep fusion, or late fusion schemes. However, their impact on downstream decision-making tasks, especially path planning, is seldom investigated. Furthermore, even when planning is considered~\cite{yu2025end, cui2022coopernaut, liu2025towards, lei2025risk, fang2025towards} experiments are often confined to simulation environments, lacking real-vehicle closed-loop validation. Thus, the practical benefits of cooperative perception for driving decisions remain largely unverified. 

In this work, we take a step forward by explicitly demonstrating how cooperative perception enhances planning performance in real vehicles. However, achieving this goal introduces three major challenges: First, ensuring each vehicle maintains accurate 3D perception and robust global localization, which are the foundations of cooperative perception; Second, designing an interface that allows cooperative perception to directly benefit conventional planning pipelines without architectural changes; Third, deploying the system under real-world communication constraints, where bandwidth and latency are critical.  

To address these challenges, we propose CooperDrive, a cooperative perception framework with three key components. First, we design a multi-task perception network that jointly performs 3D detection and semantic localization within a shared backbone, generating Bird's-Eye View (BEV) features that serve both perception and localization, ensuring accurate detection and localization for each vehicle. Second, we aggregate detection results from multi-agent into a global BEV representation that can be directly used by a standard hierarchical planner. Finally, we implement and deploy the system on real vehicles equipped with LiDAR, GPS/IMU, and prototype vehicle-to-vehicle (V2V) radios, validating its efficiency under strict bandwidth and latency conditions.  
\begin{figure*}[htbp]
    \centerline{
    \includegraphics[width=6in, height=3in]{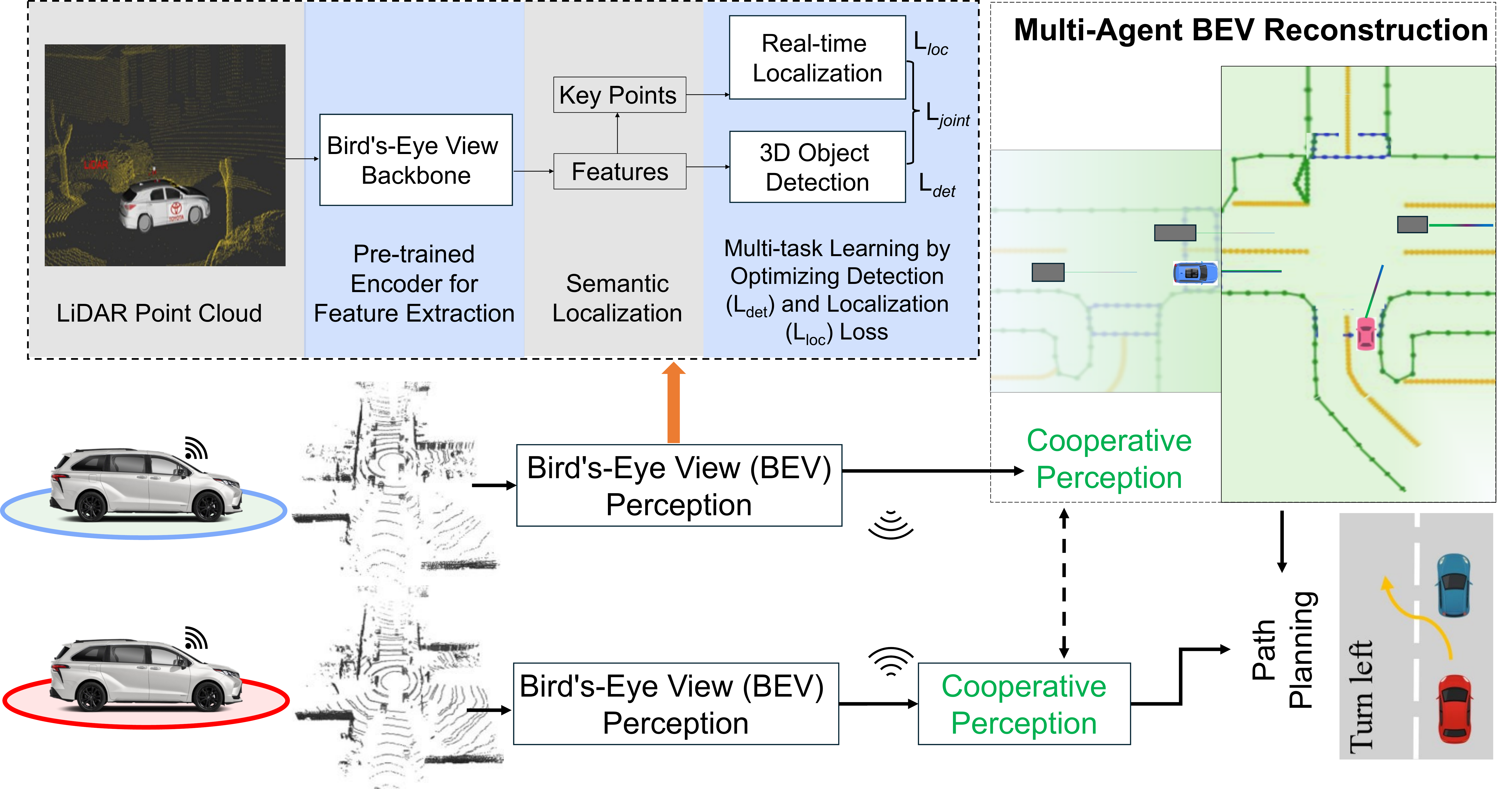}\\
    }
    \caption{Reconstructed Bird’s-Eye View (BEV) representation generated through cooperative perception, where each vehicle contributes precise localization and detection outputs from our proposed Multi-task BEV Perception Network to enhance situational awareness and support safer path planning.}
    \label{fig:2}
    \vspace{-3mm}
\end{figure*}
The contributions of this work are as follows:
\begin{itemize}
\item To the best of our knowledge, we present the first bandwidth-efficient cooperative driving system validated through real-vehicle experiments. Results demonstrate safer and more efficient decision-making in occlusion-prone scenarios.
\item We propose a novel multi-task perception network that unifies detection and semantic localization, where deeper BEV features drive accurate detection and earlier BEV features yield geometry-aware localization, mutually reinforcing each other.
\item We provide real-world evidence that cooperative perception directly enhances path planning: by sharing object-level results, vehicles anticipate hazards earlier, transforming planning from reactive to predictive without any modification to the existing planning architecture.
\end{itemize}

\section{Related Works}
 

\textbf{Autonomous driving.}
Research in autonomous driving has followed two main paradigms: modular pipelines and end-to-end learning. Modular systems such as Autoware~\cite{kato2018autoware} and Apollo~\cite{fan2018baidu} separate perception, localization, planning, and control into interpretable modules, and have been widely used in both academia and industry. In parallel, end-to-end methods have advanced rapidly, with representative BEV-based frameworks such as UniAD~\cite{hu2023planning}, VAD~\cite{jiang2023vad}, and ST-P3~\cite{hu2022st}. However, both paradigms predominantly focus on single-vehicle capabilities, overlooking the potential of cooperative mechanisms to address perception limitations in occluded or complex environments.

\textbf{Cooperative perception.}  
To address the inherent limitations of ego-centric perception under occlusion and restricted sensing range, cooperative perception has been extensively studied. Existing approaches are commonly categorized by the level of information exchange. Early-fusion methods transmit raw sensor data such as LiDAR point clouds, as in Cooper~\cite{chen2019cooper}, which can improve coverage but incur prohibitive bandwidth. Deep-fusion methods instead share intermediate BEV features, with examples including F-Cooper~\cite{chen2019f}, V2XViT~\cite{xu2022v2x}, CoBEVT~\cite{xu2022cobevt}, SiCP~\cite{qu2024sicp}, and HEAD~\cite{qu2024head}, which achieve higher accuracy by learning to align and aggregate BEV features. Late-fusion approaches only transmit object-level results such as 3D bounding boxes, which are more bandwidth-efficient and easier to integrate into existing pipelines. While these methods effectively enhance detection accuracy, most evaluations remain limited to perception benchmarks, leaving the impact of cooperative perception on downstream planning and decision making largely unexplored.

\textbf{Cooperative autonomous driving.}  
Building on cooperative perception, recent studies have explored integrating V2X into autonomous driving stacks to enable collaborative decision-making and planning. Examples include UniV2X~\cite{yu2025end}, COOPERNAUT~\cite{cui2022coopernaut}, V2Xverse~\cite{liu2025towards}, RiskMM~\cite{lei2025risk}, CoDrivingLLM~\cite{fang2025towards}, Co-MTP~\cite{zhang2025co}, CMP ~\cite{wang2025cmp}, V2XPnP~\cite{zhou2025v2xpnp} and AutowareV2X~\cite{asabe2023autowarev2x}. These studies highlight the potential of incorporating cooperative information into driving policies. However, some focus on vehicle-to-infrastructure communication and real-vehicle validation remains limited in most prior work. In contrast, our system is deployed on real vehicles, demonstrating that cooperative perception can enhance driving decisions in practice.

\section{Methodology}
Shown in Figure~\ref{fig:2}, this section presents the per-vehicle architecture, consisting of two tightly coupled components: a multi-task perception network (Section~\ref{sec:joint detection and localization}) and a hierarchical planner (Section~\ref{sec:planning}). The network performs 3D object detection and ego vehicle localization, then transforms all detections into a shared world coordinate frame. The planner then uses this structured representation, along with other onboard information, to compute safe and feasible trajectories.
%
A key advantage of this design is its native support for cooperative perception. Since all vehicles operate in the same world frame, cooperative detections can be integrated with no additional inter-vehicle coordinate transformation.

\subsection{Multi-task Perception Network}
\label{sec:joint detection and localization}
We use a bird's-eye view (BEV) 3D perception network to support two parallel heads: a CenterPoint-style detection head on deeper BEV features and a localization head on earlier BEV features. The point cloud is voxelized within a fixed range to form a BEV grid with resolution \(H \times W\), which defines the spatial support for both heads. This design provides three benefits:
(i) \emph{efficiency}, since one forward pass serves both tasks;
(ii) \emph{consistency and robustness}, as detection-derived boxes are dilated on the BEV plane to form a \emph{dynamic mask} that suppresses moving actors during localization supervision and pose estimation;
(iii) \emph{multi-task regularization}, where geometry-aware supervision steers shared features toward road layout and static discontinuities, improving generalization in sparse and long-range regimes.

\subsubsection{3D Object Detection}
\label{sec:perception}

The detection head operates on the final BEV feature map and predicts a class-wise center heatmap together with per-cell regression targets. The center heatmap is defined as
\begin{equation}
\hat{H}_t \in [0,1]^{H \times W \times C},
\end{equation}
where $C$ is the number of object classes. For each ground truth 3D box, a 2D Gaussian is rendered at the projected center on the BEV plane. The heatmap is trained with a variant of focal loss applied channel-wise, encouraging high response near object centers and suppressing low-confidence false positives.
In parallel, the regression head predicts a 10-dimensional vector at each BEV cell:
\begin{equation}
\hat{\mathbf{b}}_t = 
[\Delta x,\ \Delta y,\ z,\ l,\ w,\ h,\ \sin\theta,\ \cos\theta,\ v_x,\ v_y]^{\top},
\end{equation}
where $(\Delta x, \Delta y)$ is a sub-grid offset, $z$ is height above ground, $(l, w, h)$ are box dimensions, $(\sin\theta, \cos\theta)$ encode orientation, and $(v_x, v_y)$ denote planar velocity. These attributes are supervised with an L1 loss computed only at positive center locations.

At inference, local maxima in $\hat{H}_t$ are extracted as object centers. For each center, the corresponding regression values are gathered from $\hat{\mathbf{b}}_t$ to decode an oriented 3D box on the BEV plane and lift it to full 3D using the predicted height. A distance-based non-maximum suppression is applied to remove duplicates.
The resulting detection output at time $t$ is a set of oriented 3D bounding boxes:
\begin{equation}
\mathcal{O}_t = \left\{
\left(\mathbf{p}_j,\ \mathbf{s}_j,\ \theta_j,\ \mathbf{v}_j,\ c_j\right)
\right\}_{j=1}^{N_t},
\end{equation}
where $\mathbf{p}_j = (x_j, y_j, z_j)$ is the 3D center, $\mathbf{s}_j = (l_j, w_j, h_j)$ is the box size, $\theta_j$ is the yaw angle, $\mathbf{v}_j = (v_{x,j}, v_{y,j})$ is the planar velocity, and $c_j$ is the class label.
The final boxes are dilated on the BEV plane to produce a \emph{dynamic mask}. This mask is used during localization to suppress regions affected by moving objects in both training and pose estimation.

\subsubsection{Semantic Localization}
\label{sec:localization}

\begin{figure*}[htbp]
    \centerline{
    \includegraphics[width=6.1in, height=2.1in]{figure/fig3.pdf}\\
    }
    \caption{Comparison of a left-turn intersection scenario (real-world) under ego-only perception and CooperDrive over time (t0 to t10). 
\textbf{Top:} With ego-only perception, the oncoming vehicle from the left remains occluded and undetected until t9 when it enters the collision zone, leading to a potential collision risk. 
\textbf{Bottom:} With cooperative perception, the sender detects the dynamic vehicle from t0 and continuously shares it with the ego vehicle. This enables the ego to maintain awareness throughout the sequence and plan an early stop before the collision zone, ensuring safety.}
    \label{fig:3}
    \vspace{-3mm}
\end{figure*}

The goal of localization is to accurately estimate the vehicle pose in the world coordinate system.
Unlike conventional methods that directly match raw LiDAR point clouds to prebuilt maps, we avoid using dense point clouds due to their high storage cost. We seek a lightweight and geometrically explicit representation that is stable under dynamic traffic and efficient to store and match.
To this end, we decompose semantic localization into two steps: (i) learning a \emph{keypoint heatmap} that highlights LiDAR-observable static geometric boundaries, and (ii) performing scan-to-map pose refinement by aligning sparse keypoints to a lightweight \emph{global geometry map}.

\textbf{Keypoint heatmap learning.}
We attach a dedicated localization head to an earlier stage BEV feature map and predict a keypoint heatmap $E_t \in [0,1]^{H \times W}$, where high responses indicate persistent geometric discontinuities that are useful for localization (e.g., road boundary, sidewalk edges, and wall edges that are reliably observable in LiDAR). 

During training, we construct the ground truth binary boundary mask $Y_t^* \in \{0,1\}^{H \times W}$ from available static boundary annotations in nuScenes dataset (from map and lidarseg). We then train the heatmap using a weighted binary cross-entropy loss on BEV cells, and compute the heatmap loss only on static regions outside the dynamic mask. 

\textbf{Keypoint extraction.}
We discretize $E_t$ into a sparse keypoint set $\mathcal{K}_t$ by applying peak selection and a confidence threshold. We then discard keypoints that fall inside the dynamic mask, and the remaining static keypoints are then used for both global geometry map construction and subsequent scan-to-map pose estimation.

%
%

\textbf{Pose estimation.}
First, we construct a global geometry map in the world frame by transforming the extracted static keypoints into the world coordinate system using the current pose estimate and accumulating them over time. This map serves as a persistent geometric reference for subsequent localization and pose refinement.

Second, given a pose prior (e.g., a GNSS-initialized pose), we retrieve the corresponding local region of the global geometry map and use it to refine the current vehicle pose. The pose prior provides an initial placement of the current keypoint set in the world frame, and existing scan-to-map matching method is then applied between the current keypoints and the retrieved map region within a bounded search window. To improve robustness under partial observations and ambiguous linear boundary structures, we adopt a coarse-to-fine refinement strategy: a relaxed initial alignment is first used to obtain a stable pose update, followed by a stricter refinement stage that retains only reliable correspondences. The resulting scan-to-map matching optimization refines the current pose estimate, yielding the final 2D vehicle pose $(x,y,\theta)$ in the world frame.

Finally, we compose the refined vehicle pose with the vehicle-to-sensor extrinsic to obtain the sensor pose, which is then used to transform detected objects into the global frame for downstream planning and driving decisions.

\subsubsection{Loss and Training}
\label{sec:loss-training}

We train the network with joint supervision for detection and localization:
\begin{equation}
\mathcal{L}_{\text{joint}} = \mathcal{L}_{\text{det}} + \lambda_{\text{loc}} \cdot \mathcal{L}_{\text{loc}},
\end{equation}
where $\lambda_{\text{loc}}$ balances the two objectives.

\textbf{Detection loss.}
The detection loss includes a focal loss on the center heatmap $\hat{H}_t$ and an L1 regression loss on box parameters $\hat{\mathbf{b}}_t$:
\begin{equation}
\mathcal{L}_{\text{det}} = \text{Focal}(\hat{H}_t, H_t^*) \;+\; \lambda_{\text{reg}} \cdot \left\| \hat{\mathbf{b}}_t - \mathbf{b}_t^* \right\|_1,
\end{equation}
where $H_t^*$ is the ground truth heatmap and $\mathbf{b}_t^*$ denotes the ground truth box regression targets. $\lambda_{\text{reg}}$ is a weighting term.

\textbf{Localization loss.}
The localization head predicts a keypoint heatmap $E_t \in [0,1]^{H \times W}$, supervised by a binary boundary mask $Y_t^* \in \{0,1\}^{H \times W}$. Since boundary cells are extremely sparse, we use the Weighted Binary Cross-Entropy (WBCE) loss, where positive boundary cells are up-weighted to address severe class imbalance:
\begin{equation}
\mathcal{L}_{\text{loc}} = \text{WBCE}(E_t, Y_t^*),
\end{equation}
We compute this loss on static regions not covered by the dynamic mask, so that localization relies on static structures.

\textbf{Training strategy.}
We adopt a staged training procedure. First, we train the backbone and detection head to convergence. Then, we freeze them and train the localization head. Finally, we fine-tune all components jointly to improve overall consistency.

\subsection{Path Planning}
\label{sec:planning}
%
We adopt a conventional hierarchical path planning framework, comprising behavior planning for high-level decisions and motion planning for trajectory refinement. Behavior planning operates over a longer horizon to select a reference path and a velocity envelope, while motion planning runs on a shorter horizon to generate a dynamically feasible and smooth trajectory. The planning framework remains unchanged, generating safe and efficient trajectories for the ego vehicle by integrating an extended set of object detections from cooperative vehicles. We demonstrate that, compared to single-vehicle perception, the enriched perceptual input enables the planner to anticipate potential conflicts earlier and produce safer trajectories.

The planning process iteratively refines trajectories based on fused sensor inputs, the output is a time-parameterized trajectory over a finite horizon. Perception fusion assigns higher weights to ego observations when the object is within the field of view, while remote detections are down-weighted by latency, confidence, and source trust; this yields the fused set \(\mathcal{O}_{\text{fused}} = \mathcal{O}_{\text{ego}} \cup \mathcal{O}_{\text{Coop}}\).

\subsubsection{Behavior Planning}
Behavior planning generates a reference path with associated velocity constraints by separating lateral and longitudinal decisions for efficiency.

First, the lateral path is constructed along the global route, starting from a centerline reference extracted from the map. To accommodate maneuvers like lane changes or obstacle avoidance, the path is laterally shifted using a constant-jerk profile. Let \(s\) denote the arc-length along the reference centerline. The lateral offset \(y(s)\) is modeled as a piecewise cubic polynomial (spline) to ensure smooth transitions:
\begin{equation}
y(s) = 
\begin{cases} 
a_1 s^3 + b_1 s^2 + c_1 s + d_1, & 0 \leq s < s_1 \\
a_2 s^3 + b_2 s^2 + c_2 s + d_2, & s_1 \leq s < s_2 \\
\vdots & \vdots
\end{cases}
\label{eq:1}
\end{equation}
Coefficients \(a_i, b_i, c_i, d_i\) are solved such that the path starts and ends smoothly, with continuous position, speed, and acceleration where segments meet. Candidate paths are sampled by varying shift magnitudes and evaluated via a cost function:
\begin{equation}
\begin{aligned}
C ={}&\, w_d \int_0^S (y(s) - y_{\text{ref}}(s))^2\, ds \\
&\ + w_c \int_0^S \sum_{o \in \mathcal{O}_{\text{fused}}}
      \max\!\bigl(0,\, d_{\min} - \phi_o(s)\bigr)\, ds ,
\end{aligned}
\label{eq:2}
\end{equation}

\noindent where \(\phi_o(s) \triangleq \min_{\tau \in [0,T_{\text{pred}}]}
d\!\bigl(p_\gamma(s),\, \mathcal{B}_o(\tau)\bigr)\).
Here, \(p_\gamma(s)=c(s)+n(s)\,y(s)\) denotes the path point in the world frame
(with \(c(s)\) the centerline and \(n(s)\) its unit normal),
\(\mathcal{B}_o(\tau)\) denotes the predicted footprint of object \(o\) at future time \(\tau\),
and \(d(\cdot,\cdot)\) is the Euclidean point-to-footprint distance.
\(w_d,w_c\) are weights for path deviation and collision risk, \(\mathcal{O}_{\text{fused}}\) is the fused object set,
\(d_{\min}\) is the safety margin, and \(T_{\text{pred}}\) is the prediction horizon. The set \(\mathcal{O}_{\text{fused}}\) includes occluded objects contributed by cooperative perception, thereby triggering earlier penalties even when ego-only perception blind.

Cooperative perception advantages manifest in the lateral path: Fused detections from remote vehicles extend \(\mathcal{O}_{\text{fused}}\) to include occluded objects (e.g., cross-traffic at intersections), triggering earlier path shifts or lane changes. In experiments, this reduced collision risk through proactive avoidance compared to ego perception alone.

Next, longitudinal velocity constraints are synthesized in a scene-aware manner, conditioned on map semantics and multi-object predictions (e.g., intersections or crosswalks). For each segment, deceleration or stop points are inserted based on predicted obstacle trajectories. The maximum velocity profile \(v_{\text{max}}(s)\) is computed as:
\begin{equation}
v_{\text{max}}(s)
 = \min\bigl( v_{\text{map}}(s),\, v_{\text{stop}}(s),\, v_{\kappa}(s) \bigr).
\label{eq:3}
\end{equation}

\noindent
where \(v_{\text{map}}(s)\) is the map-imposed limit, 
\(a_{\text{max}}>0\) is the deceleration bound, 
\(v_{\text{stop}}(s)=\sqrt{\,2\,a_{\text{max}}\,\Delta s_{\text{stop}}(s, \mathcal{O}_{\text{fused}})\,}\),
\(v_{\kappa}(s)=\sqrt{\,a_{\text{lat,max}}/\lvert \kappa(s) \rvert\,}\), and
\(\Delta s_{\text{stop}}(s, \mathcal{O}_{\text{fused}})=
\min\!\bigl\{(s_{\text{stopline}}-s)_+,\, \min_{o\in \mathcal{O}_{\text{fused}}}(s_{\text{conf},o}-s)_+\bigr\}\)
is the arc-length to the nearest conflict.
Here \(a_{\text{lat,max}}\) is the lateral-acceleration bound,
\((x)_+=\max(x,0)\), \(\mu\) is the tire--road friction coefficient,
\(g\) is gravitational acceleration, and \(\kappa(s)\) is curvature.
By construction, the cooperative envelope \(v_{\text{max}}^{\text{coop}}(s):=v_{\text{max}}(s;\mathcal{O}_{\text{fused}})\)
satisfies \(v_{\text{max}}^{\text{coop}}(s)\le v_{\text{max}}(s;\mathcal{O}_{\text{ego}})\) for all \(s\).

Cooperative perception fusion augments predictions by incorporating remote confidences. This enables earlier velocity reductions, e.g., preempting stops for unseen vehicles, improving safety margins in intersection tests.

\subsubsection{Motion Planning}
Motion planning refines the behavioral path into a feasible trajectory by separating lateral optimization and longitudinal smoothing.

First, the lateral optimization adjusts positions within the drivable area using quadratic programming (QP) to minimize deviations while avoiding obstacles, with non-linear distances linearized around the current path and updated iteratively:
\begin{equation}
\begin{aligned}
\min_{\mathbf{y}}\quad
& \mathbf{y}^{\top} Q \mathbf{y} + \mathbf{c}^{\top} \mathbf{y} \\
\text{s.t.}\quad
& \mathbf{A}\!\big(\mathbf{y}^{\mathrm{ref}};\, \mathcal{O}_{\text{fused}}\big)\,\mathbf{y}
  \le \mathbf{b}\!\big(\mathbf{y}^{\mathrm{ref}};\, \mathcal{O}_{\text{fused}}\big), \\
& \mathbf{y}_{\min} \le \mathbf{y} \le \mathbf{y}_{\max}.
\end{aligned}
\label{eq:4}
\end{equation}
\noindent
Here \(\mathbf{y}=[y(s_1),\ldots,y(s_{N_s})]^{\top}\) are lateral offsets at sampled arc-length points,
\(Q\succeq 0\) penalizes smoothness,
\(\mathbf{c}\) encourages adherence to the reference path, and
\(\mathbf{A}(\cdot;\mathcal{O}_{\text{fused}}),\mathbf{b}(\cdot;\mathcal{O}_{\text{fused}})\) are assembled from
linearized signed–distance constraints to the predicted footprints of all objects in \(\mathcal{O}_{\text{fused}}\)
and are relinearized each cycle.

The per-step limits \(v_{\text{max},k}\) are obtained by sampling the cooperative envelope
\(v_{\text{max}}^{\text{coop}}(s)\), computed with \(\mathcal{O}_{\text{fused}}\), along the reference path.
Next, the longitudinal smoothing optimizes motion via a jerk-constrained convex quadratic program (QP) to produce a smooth velocity profile \(v(t)\):
\begin{equation}
\begin{aligned}
&\min \sum_{k=1}^{K-1} j_k^2 \;+\; \epsilon \sum_{k=1}^{K} \bigl(v_k-v_{\text{max},k}\bigr)^2 \\
\text{s.t. }\;& v_{k+1}=v_k+a_k\Delta t,\quad a_{k+1}=a_k+j_k\Delta t,\\
& v_{\min}\le v_k\le v_{\text{max},k},\ \ |a_k|\le a_{\text{max}},\ \ |j_k|\le j_{\text{max}},\\
& v_k=0\ \text{for } k\ge k_{\text{stop}}.
\end{aligned}
\label{eq:6}
\end{equation}

Here \(k = 1, \dots, K\) indexes discrete time steps over the horizon; \(v_k, a_k, j_k\) are velocity, acceleration, and jerk at step \(k\); \(\epsilon\) balances adherence to the maximum velocity limits \(v_{\text{max},k}\) from upstream; and the constraints propagate vehicle dynamics while bounding states for feasibility and comfort. Here \(k_{\text{stop}}\) is the first time index at or beyond the virtual stop line projected onto the trajectory grid. Cooperative perception influences this indirectly via upstream \(v_{\text{max},k}\) reductions from fused detections, enabling rapid yet comfortable responses, e.g., earlier braking for cross-traffic, reducing peak decelerations.

\subsubsection{Cooperative perception enhanced path planning}
Cooperative perception turns reactive planning into predictive by extending the ego’s awareness with remote detections. 
As shown in Figure~\ref{fig:3}, in a real-world left-turn intersection, ego-only perception fails to detect the oncoming vehicle until it enters the collision zone. 
With cooperative perception, the sender shares the vehicle’s trajectory early, allowing the ego to adjust its plan in advance and execute a safer stop.

\section{Experiments}

We design experiments at three complementary levels. First, we evaluate detection and localization quantitatively on the nuScenes dataset ~\cite{caesar2020nuscenes} to assess accuracy and robustness under controlled drift conditions. Second, we deploy our cooperative perception system on real vehicles to demonstrate its impact on downstream planning in occlusion-prone urban scenarios. Finally, we analyze communication efficiency and real-time feasibility to verify that the proposed framework can operate reliably in real vehicles.


\subsection{Experiment Setup}
%

\textbf{Dataset.} We use the nuScenes dataset~\cite{caesar2020nuscenes} for detection and localization evaluation. For 3D object detection, we follow the official train/validation split. 
%
For localization, we leverage the accurate ground truth ego poses provided by nuScenes. Following~\cite{dong2023lidar}, we simulate a noisy GNSS prior by injecting zero-mean Gaussian perturbations into the initial 2D pose:
$\Delta x,\Delta y \sim \mathcal{N}(0,(\alpha\sigma_{xy})^2)$ and
$\Delta\theta \sim \mathcal{N}(0,(\alpha\sigma_{\theta})^2)$,
where $\sigma_{xy}=1$ m, $\sigma_{\theta}=2^\circ$, and $\alpha \in \{1,2,3,4\}$ denotes the error level. For each error level, we report the mean translation and heading errors over multiple random trials.

\textbf{Real-vehicle platform.} Two Toyota Sienna vehicles are equipped with an Ouster LiDAR, GPS/IMU, and prototype V2V radios. Each runs a ROS2-based pipeline on an NVIDIA GPU with 16 GB VRAM. The ego and sender vehicles perform detection and localization independently, and share detections over real-time V2V communication to improve vehicle planning.

\begin{table}[t]
\centering
\caption{Localization evaluation under noisy GNSS initialization on the nuScenes dataset. We compare the noisy GNSS prior with ICP and NDT scan-to-map pose refinement using the learned current keypoints and the global geometry map.}
\label{tab:loc_table}
\setlength{\tabcolsep}{4pt}
\renewcommand{\arraystretch}{1.05}
\begin{tabular}{c|ccc|ccc}
\toprule
\multirow{2}{*}{\textbf{Error Level}} &
\multicolumn{3}{c|}{\textbf{Translation error (m)$\downarrow$}} &
\multicolumn{3}{c}{\textbf{Heading error (deg)$\downarrow$}} \\
& GNSS & ICP & NDT & GNSS & ICP & NDT \\
\midrule
1 & 2.26 & 1.57 & 0.52 & 4.47 & 1.26 & 0.74 \\
2 & 3.41 & 1.96 & 0.71 & 6.68 & 1.53 & 0.96 \\
3 & 4.51 & 2.76 & 1.32 & 8.97 & 4.53 & 1.90 \\
4 & 5.61 & 4.03 & 2.12 & 11.26 & 8.24 & 2.82 \\
\bottomrule
\end{tabular}
\end{table}

\begin{table}[t]
\centering
\caption{LiDAR 3D object detection on nuScenes dataset.}
\label{tab:perception_nus}
\setlength{\tabcolsep}{8pt}
\renewcommand{\arraystretch}{1.12}
\begin{tabular}{lccc}
\toprule
Model & Backbone & mAP$\uparrow$ & NDS$\uparrow$ \\
\midrule
SSN ~\cite{zhu2020ssn}& SECFPN & 35.17 & 49.76 \\
PointPillars ~\cite{lang2019pointpillars} & SECFPN & 35.19 & 50.27 \\
CenterPoint ~\cite{yin2021center}& SECFPN & 56.11 & 64.61 \\
\textbf{Ours} & SECFPN & \textbf{56.27} & \textbf{64.78} \\
\bottomrule
\end{tabular}
\end{table}


\begin{table}[t]
\centering
\setlength{\tabcolsep}{2.8pt}
\renewcommand{\arraystretch}{1.11}
\caption{Overall planning metrics comparison across all test scenarios: Ego-only vs. CooperDrive.}
\label{tab:planning_table}
\begin{tabular}{lcccc}
\toprule
Method & $\mathrm{TTC}_{\min}$ (s)$\uparrow$ & $\mathrm{DRAC}$ (m/s$^2$)$\downarrow$  & $\mathrm{DCZ}$ (m)$\uparrow$ & VR (\%)$\downarrow$ \\
\midrule
Ego-only          & $2.05$ & $0.312$ & $2.01$ & $18$ \\
CooperDrive  & $6.62$ & $0.060$ & $4.12$ & $2$ \\
\midrule
\textbf{Improvement}       & $\textbf{+4.57}$ & $\textbf{-0.252}$ & $\textbf{+2.11}$ & $\textbf{-16}$ \\
\bottomrule
\end{tabular}
\end{table}
\textbf{Baselines.} For detection, we compare with SSN~\cite{zhu2020ssn}, PointPillars~\cite{lang2019pointpillars}, and CenterPoint~\cite{yin2021center}. We follow the corresponding MMDetection3D~\cite{mmdet3d2020} implementations, and use a unified SECFPN backbone across all methods for a fair comparison.
%
For localization, we use the noisy GNSS pose as the initial estimate, and adopt ICP~\cite{besl1992method} and NDT~\cite{biber2003normal} as representative scan-to-map refinement baselines. Since both already provide strong localization performance, they are sufficient for evaluating the effectiveness of the learned keypoints.
For planning, we compare ego-only planning against cooperative perception, where the planner is unchanged but receives additional detections from the collaborating vehicle.
We conduct planning tests across multiple scenarios in a complex urban environment, covering both Line-of-Sight (LOS) and Non-Line-of-Sight (NLOS) conditions.

%
%
\textbf{Metrics.} Detection is evaluated using \textit{mean Average Precision (mAP)} and \textit{nuScenes Detection Score (NDS)} \cite{caesar2020nuscenes}, while localization is measured by \textit{translation error} (meters) and \textit{heading error} (degrees).
Planning is assessed with standard metrics widely used in autonomous driving. \textit{Minimum Time-to-Collision ($\mathrm{TTC}_{\min}$)} indicates the shortest predicted time to a potential collision, with larger values reflecting safer planning \cite{sultan2003assessing}. \textit{Deceleration Rate to Avoid Crash (DRAC)} measures the minimum deceleration required to avoid collisions, with lower values indicating safer and smoother trajectories \cite{westhofen2021criticality}. \textit{Distance to Conflict Zone (DCZ)} is the shortest distance to a predicted conflict zone, representing potential collision regions, with larger distances indicating safer paths \cite{shetty2021safety}. \textit{Violation Rate (VR)} quantifies the proportion of planning steps violating safety constraints \cite{wang2023does}, primarily entering conflict zones or violating minimum safety distances, serving as a surrogate for \textit{Collision Rate (CR)} in the absence of real collisions.

\subsection{Robust Detection and Localization}
%
Table~\ref{tab:loc_table} reports localization performance under injected GNSS pose drift. Both ICP and NDT refine the noisy GNSS prior by aligning the learned current frame keypoints to the global geometry map, substantially reducing both translation and heading errors. NDT consistently outperforms ICP across all error levels, indicating stronger robustness under noisy initialization. ICP also improves over the GNSS prior, showing that the learned keypoints preserve meaningful geometric structure for scan-to-map alignment. As NDT is more stable under larger initialization errors, we use it in the real-vehicle experiments. Overall, these results show that the learned keypoints provide reliable geometric support for pose refinement across different methods and noise levels.

Table~\ref{tab:perception_nus} shows 3D object detection performance. 
Our method even outperforms CenterPoint, achieving the highest mAP and NDS on the nuScenes dataset. 
This shows that our joint design does not sacrifice detection quality; rather, it further enhances accuracy compared to baselines.

The reason why our approach excels in both detection and localization is the shared backbone design: 
geometry aware BEV features learned for detection provide strong structural cues that also benefit localization, while the localization head enforces geometric consistency that regularizes the feature space for detection. This mutual reinforcement explains why the two tasks improve together in our unified framework.

\subsection{Enhanced Path Planning}
We now examine how cooperative perception improves planning when integrated into a conventional hierarchical planner.
Quantitative results in Table~\ref{tab:planning_table} highlight consistent advantages. Compared to ego-only inputs, CooperDrive increases the minimum time-to-collision by more than $223\%$, reduces the deceleration rate to avoid crash by a factor of five, and enlarges the distance to conflict-zone. Moreover, the violation rate drops from 18\% to 2\%, indicating that the planner produces more predictable and less aggressive maneuvers once occluded vehicles are revealed. These improvements arise because the fused set $\mathcal{O}_{\text{fused}}$ includes vehicles that would otherwise remain hidden, triggering earlier velocity reductions and safer trajectory adjustments.
\begin{figure*}[!htp]
  \begin{center}
  \includegraphics[width=7in, height=1.9in]{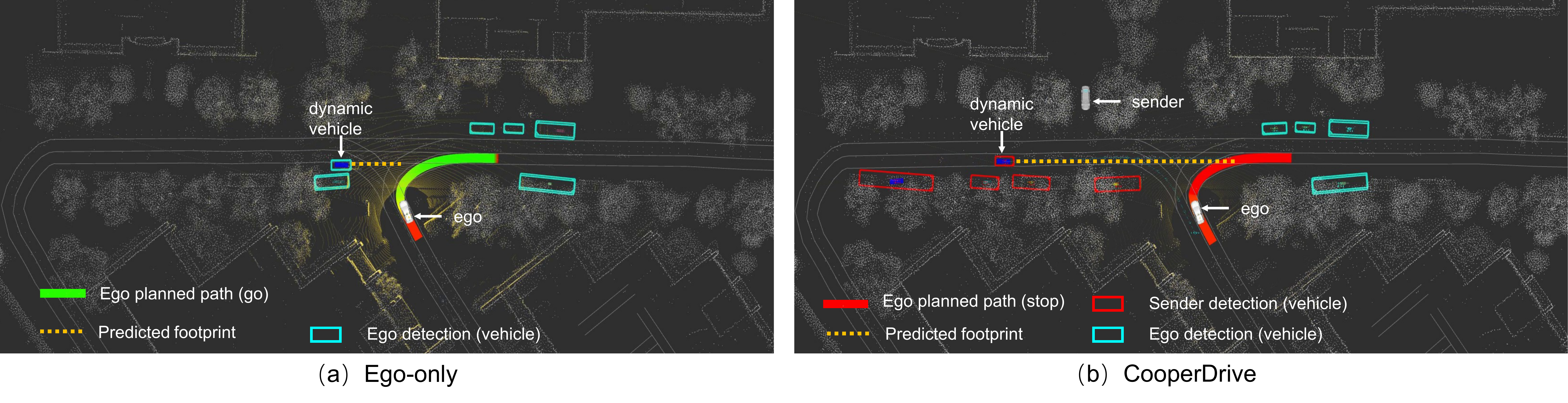}\\
   \caption{Comparison at a T-intersection right turn: Ego-only vs. CooperDrive.}
    \label{fig:5}
  \end{center}
  \vspace{-3mm}
\end{figure*}

\begin{figure*}[!htp]
  \begin{center}
  \includegraphics[width=7.2in, height=1in]{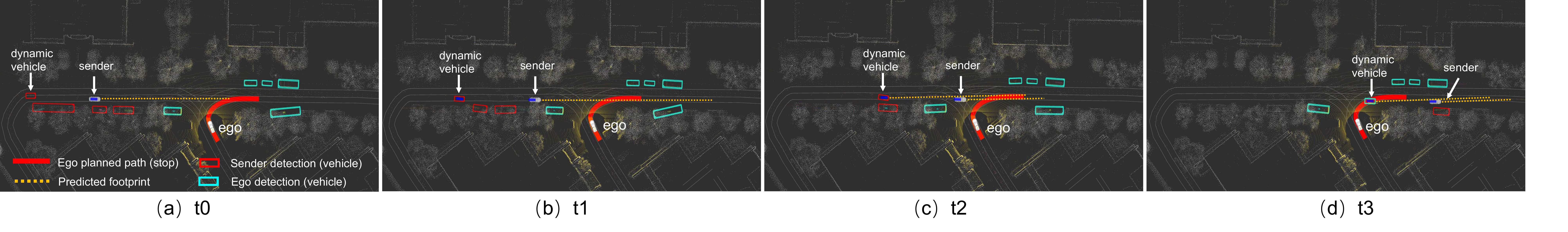}\\
   \caption{CooperDrive at a T-intersection right turn with a sender emerging from the ego's blind spot.}

    \label{fig:6}
  \end{center}
  \vspace{-2mm}
\end{figure*}

\subsection{Real-world Deployment}
\begin{figure}[t]
\centering
\includegraphics[width=1\linewidth]{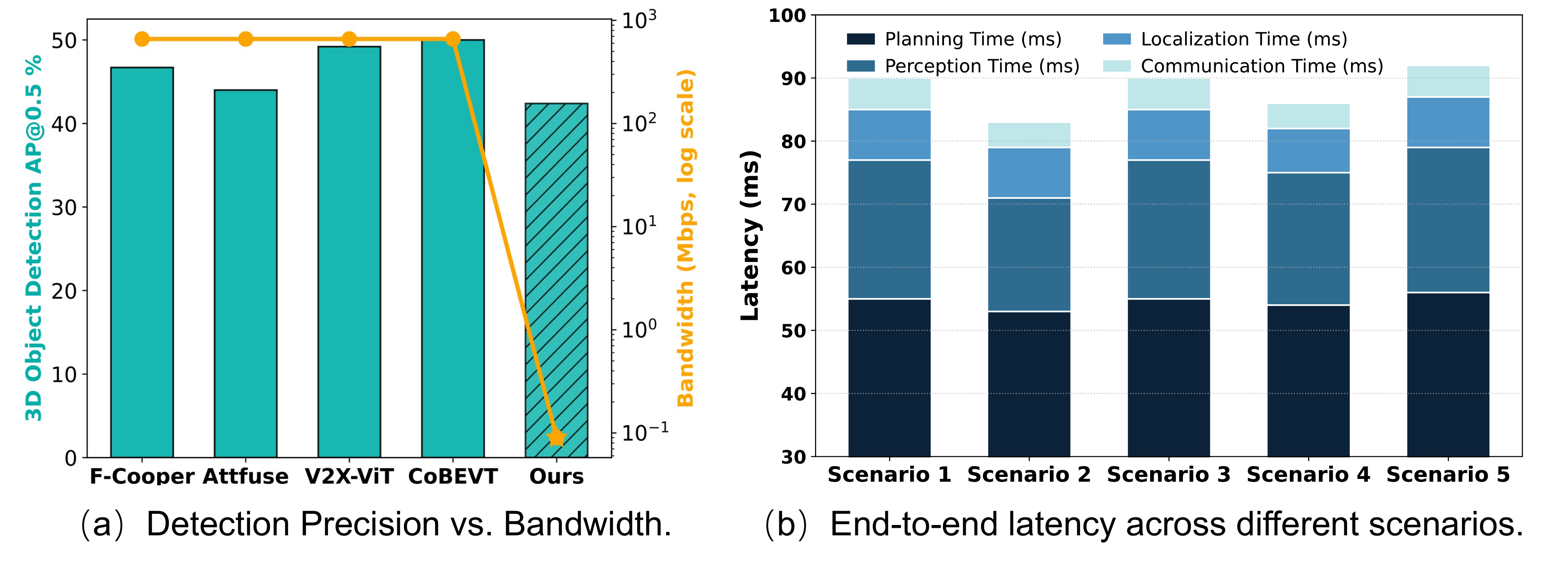}
\caption{Real-world deployment analysis of communication efficiency and end-to-end latency.}
\label{fig:bandwidth}
\vspace{-4mm}
\end{figure}
Qualitative cases further illustrate this effect. In Figure~\ref{fig:5}, at a T-intersection, the ego-only planner fails to detect the oncoming vehicle until it enters the collision zone, leading to a late and abrupt stop. With cooperative perception, the sender shares the trajectory of the occluded vehicle from the start, allowing the ego to anticipate the conflict and plan a smooth, early stop. In Figure~\ref{fig:6}, another sender emerges from the ego’s blind spot, followed by an additional vehicle. Cooperative perception ensures that both are continuously tracked through shared detections, enabling proactive planning and maintaining stable communication. Together, these results demonstrate that cooperative perception transforms planning behavior from reactive to predictive, enhancing both safety margins and driving comfort in occlusion-prone scenarios.

We first clarify why we do not adopt BEV feature-level fusion. As shown in Figures~\ref{fig:bandwidth} (a), and based on our prior study in HEAD~\cite{qu2024head} using the V2V4Real dataset~\cite{xu2023v2v4real}, feature-sharing methods such as F-Cooper, AttFuse, V2X-ViT, and CoBEVT require over 660\,\text{Mbps} of communication bandwidth. In contrast, our method transmits only compact object-level outputs at 0.09\,\text{Mbps}. While feature-sharing strategies may achieve slightly higher detection accuracy, the communication overhead is more than three orders of magnitude larger. Given the constrained bandwidth in our vehicular networks, these methods are not feasible for our real-time deployment.

To further verify practical feasibility, we measure the end-to-end latency from sensing to planning across five representative scenarios: Intersection Left Turn, Intersection Right Turn, T-intersection Left Turn, T-intersection Right Turn, and Low-visibility Intersection Crossing. As illustrated in Figure~\ref{fig:bandwidth} (b), the total latency (including perception, localization, communication, and planning) maintains an average of 89\,\text{ms}. This consistent performance confirms the system's readiness for real-time, safety-critical driving.

Beyond efficiency, it is important to clarify what information is actually exchanged. In our system, vehicles transmit their precise poses and object-level detection results. This shared data provides a richer and more reliable view of the environment, allowing each ego vehicle to anticipate hazards earlier and make safer path planning decisions.  

Finally, the benefits of V2X communication extend beyond direct V2V links. Direct communication is essential for low-latency cooperation, as in our deployment, but V2N2V links (e.g., 5G uplink/downlink) can further improve coverage and connectivity in urban areas. Moreover, V2N also supports large-scale data collection in offline mode, which is crucial for training and optimizing AI perception models.  

Since only object-level results are transmitted, our approach remains agnostic to sensor modality and backbone architecture. This enables seamless integration across heterogeneous fleets equipped with different LiDAR or Camera-based detection systems, making cooperative perception widely deployable in practice.

\section{Conclusions}
\label{sec:conclusion}
%
In this paper, we proposed CooperDrive, a cooperative perception framework designed to enhance driving decisions under realistic constraints. We introduced a per-vehicle joint detection and localization framework that aligns observations in a global coordinate frame and demonstrated that a reconstructed BEV representation is sufficient to support downstream planning. Through real-world closed-loop trials, we showed that CooperDrive enables earlier anticipation, wider safety margins, and reduced interventions, while maintaining planner consistency. Importantly, these benefits were achieved with minimal communication overhead—only kilobytes per second—and with an average end-to-end latency of 89 ms. These achievements validate that cooperative perception is both effective and practical, offering a scalable path toward safer and more predictive autonomous driving.
\bibliographystyle{IEEEtran}
\bibliography{references}

@inproceedings{kato2018autoware,
  title={Autoware on board: Enabling autonomous vehicles with embedded systems},
  author={Kato, Shinpei and Tokunaga, Shota and Maruyama, Yuya and Maeda, Seiya and Hirabayashi, Manato and Kitsukawa, Yuki and Monrroy, Abraham and Ando, Tomohito and Fujii, Yusuke and Azumi, Takuya},
  booktitle={2018 ACM/IEEE 9th International Conference on Cyber-Physical Systems (ICCPS)},
  pages={287--296},
  year={2018},
  organization={IEEE}
}

@article{fan2018baidu,
  title={Baidu apollo em motion planner},
  author={Fan, Haoyang and Zhu, Fan and Liu, Changchun and Zhang, Liangliang and Zhuang, Li and Li, Dong and Zhu, Weicheng and Hu, Jiangtao and Li, Hongye and Kong, Qi},
  journal={arXiv preprint arXiv:1807.08048},
  year={2018}
}

@inproceedings{hu2023planning,
  title={Planning-oriented autonomous driving},
  author={Hu, Yihan and Yang, Jiazhi and Chen, Li and Li, Keyu and Sima, Chonghao and Zhu, Xizhou and Chai, Siqi and Du, Senyao and Lin, Tianwei and Wang, Wenhai and others},
  booktitle={Proceedings of the IEEE/CVF conference on computer vision and pattern recognition},
  pages={17853--17862},
  year={2023}
}

@inproceedings{jiang2023vad,
  title={Vad: Vectorized scene representation for efficient autonomous driving},
  author={Jiang, Bo and Chen, Shaoyu and Xu, Qing and Liao, Bencheng and Chen, Jiajie and Zhou, Helong and Zhang, Qian and Liu, Wenyu and Huang, Chang and Wang, Xinggang},
  booktitle={Proceedings of the IEEE/CVF International Conference on Computer Vision},
  pages={8340--8350},
  year={2023}
}

@inproceedings{hu2022st,
  title={St-p3: End-to-end vision-based autonomous driving via spatial-temporal feature learning},
  author={Hu, Shengchao and Chen, Li and Wu, Penghao and Li, Hongyang and Yan, Junchi and Tao, Dacheng},
  booktitle={European Conference on Computer Vision},
  pages={533--549},
  year={2022},
  organization={Springer}
}

@inproceedings{chen2019cooper,
  title={Cooper: Cooperative perception for connected autonomous vehicles based on 3d point clouds},
  author={Chen, Qi and Tang, Sihai and Yang, Qing and Fu, Song},
  booktitle={2019 IEEE 39th International Conference on Distributed Computing Systems (ICDCS)},
  pages={514--524},
  year={2019},
  organization={IEEE}
}

@inproceedings{chen2019f,
  title={F-cooper: Feature based cooperative perception for autonomous vehicle edge computing system using 3D point clouds},
  author={Chen, Qi and Ma, Xu and Tang, Sihai and Guo, Jingda and Yang, Qing and Fu, Song},
  booktitle={Proceedings of the 4th ACM/IEEE Symposium on Edge Computing},
  pages={88--100},
  year={2019}
}

@inproceedings{xu2022v2x,
  title={V2x-vit: Vehicle-to-everything cooperative perception with vision transformer},
  author={Xu, Runsheng and Xiang, Hao and Tu, Zhengzhong and Xia, Xin and Yang, Ming-Hsuan and Ma, Jiaqi},
  booktitle={European conference on computer vision},
  pages={107--124},
  year={2022},
  organization={Springer}
}

@article{xu2022cobevt,
  title={CoBEVT: Cooperative bird's eye view semantic segmentation with sparse transformers},
  author={Xu, Runsheng and Tu, Zhengzhong and Xiang, Hao and Shao, Wei and Zhou, Bolei and Ma, Jiaqi},
  journal={arXiv preprint arXiv:2207.02202},
  year={2022}
}

@inproceedings{qu2024sicp,
  title={Sicp: Simultaneous individual and cooperative perception for 3d object detection in connected and automated vehicles},
  author={Qu, Deyuan and Chen, Qi and Bai, Tianyu and Lu, Hongsheng and Fan, Heng and Zhang, Hao and Fu, Song and Yang, Qing},
  booktitle={2024 IEEE/RSJ International Conference on Intelligent Robots and Systems (IROS)},
  pages={8905--8912},
  year={2024},
  organization={IEEE}
}

@inproceedings{qu2024head,
  title={HEAD: A Bandwidth-Efficient Cooperative Perception Approach for Heterogeneous Connected and Autonomous Vehicles},
  author={Qu, Deyuan and Chen, Qi and Zhu, Yongqi and Zhu, Yihao and Avedisov, Sergei S and Fu, Song and Yang, Qing},
  booktitle={European Conference on Computer Vision},
  pages={198--211},
  year={2024},
  organization={Springer}
}

@inproceedings{yu2025end,
  title={End-to-end autonomous driving through v2x cooperation},
  author={Yu, Haibao and Yang, Wenxian and Zhong, Jiaru and Yang, Zhenwei and Fan, Siqi and Luo, Ping and Nie, Zaiqing},
  booktitle={Proceedings of the AAAI Conference on Artificial Intelligence},
  volume={39},
  number={9},
  pages={9598--9606},
  year={2025}
}

@inproceedings{asabe2023autowarev2x,
  title={Autowarev2x: Reliable v2x communication and collective perception for autonomous driving},
  author={Asabe, Yu and Javanmardi, Ehsan and Nakazato, Jin and Tsukada, Manabu and Esaki, Hiroshi},
  booktitle={2023 IEEE 97th Vehicular Technology Conference (VTC2023-Spring)},
  pages={1--7},
  year={2023},
  organization={IEEE}
}

@article{fang2025towards,
  title={Towards interactive and learnable cooperative driving automation: a large language model-driven decision-making framework},
  author={Fang, Shiyu and Liu, Jiaqi and Ding, Mingyu and Cui, Yiming and Lv, Chen and Hang, Peng and Sun, Jian},
  journal={IEEE Transactions on Vehicular Technology},
  year={2025},
  publisher={IEEE}
}

@inproceedings{cui2022coopernaut,
  title={Coopernaut: End-to-end driving with cooperative perception for networked vehicles},
  author={Cui, Jiaxun and Qiu, Hang and Chen, Dian and Stone, Peter and Zhu, Yuke},
  booktitle={Proceedings of the IEEE/CVF Conference on Computer Vision and Pattern Recognition},
  pages={17252--17262},
  year={2022}
}

@article{liu2025towards,
  title={Towards collaborative autonomous driving: Simulation platform and end-to-end system},
  author={Liu, Genjia and Hu, Yue and Xu, Chenxin and Mao, Weibo and Ge, Junhao and Huang, Zhengxiang and Lu, Yifan and Xu, Yinda and Xia, Junkai and Wang, Yafei and others},
  journal={IEEE Transactions on Pattern Analysis and Machine Intelligence},
  year={2025},
  publisher={IEEE}
}

@article{zhang2025co,
  title={Co-mtp: A cooperative trajectory prediction framework with multi-temporal fusion for autonomous driving},
  author={Zhang, Xinyu and Zhou, Zewei and Wang, Zhaoyi and Ji, Yangjie and Huang, Yanjun and Chen, Hong},
  journal={arXiv preprint arXiv:2502.16589},
  year={2025}
}

@article{wang2025cmp,
  title={Cmp: Cooperative motion prediction with multi-agent communication},
  author={Wang, Zehao and Wang, Yuping and Wu, Zhuoyuan and Ma, Hengbo and Li, Zhaowei and Qiu, Hang and Li, Jiachen},
  journal={IEEE Robotics and Automation Letters},
  year={2025},
  publisher={IEEE}
}

@inproceedings{zhou2025v2xpnp,
  title={V2xpnp: Vehicle-to-everything spatio-temporal fusion for multi-agent perception and prediction},
  author={Zhou, Zewei and Xiang, Hao and Zheng, Zhaoliang and Zhao, Seth Z and Lei, Mingyue and Zhang, Yun and Cai, Tianhui and Liu, Xinyi and Liu, Johnson and Bajji, Maheswari and others},
  booktitle={Proceedings of the IEEE/CVF International Conference on Computer Vision},
  pages={25399--25409},
  year={2025}
}

@article{lei2025risk,
  title={Risk Map as Middleware: Toward Interpretable Cooperative End-to-End Autonomous Driving for Risk-Aware Planning},
  author={Lei, Mingyue and Zhou, Zewei and Li, Hongchen and Ma, Jiaqi and Hu, Jia},
  journal={IEEE Robotics and Automation Letters},
  volume={11},
  number={1},
  pages={818--825},
  year={2025},
  publisher={IEEE}
}

@inproceedings{caesar2020nuscenes,
  title={nuscenes: A multimodal dataset for autonomous driving},
  author={Caesar, Holger and Bankiti, Varun and Lang, Alex H and Vora, Sourabh and Liong, Venice Erin and Xu, Qiang and Krishnan, Anush and Pan, Yu and Baldan, Giancarlo and Beijbom, Oscar},
  booktitle={Proceedings of the IEEE/CVF conference on computer vision and pattern recognition},
  pages={11621--11631},
  year={2020}
}

@inproceedings{xu2023v2v4real,
  title={V2v4real: A real-world large-scale dataset for vehicle-to-vehicle cooperative perception},
  author={Xu, Runsheng and Xia, Xin and Li, Jinlong and Li, Hanzhao and Zhang, Shuo and Tu, Zhengzhong and Meng, Zonglin and Xiang, Hao and Dong, Xiaoyu and Song, Rui and others},
  booktitle={Proceedings of the IEEE/CVF conference on computer vision and pattern recognition},
  pages={13712--13722},
  year={2023}
}

@inproceedings{zhu2020ssn,
  title={Ssn: Shape signature networks for multi-class object detection from point clouds},
  author={Zhu, Xinge and Ma, Yuexin and Wang, Tai and Xu, Yan and Shi, Jianping and Lin, Dahua},
  booktitle={European Conference on Computer Vision},
  pages={581--597},
  year={2020},
  organization={Springer}
}

@inproceedings{lang2019pointpillars,
  title={Pointpillars: Fast encoders for object detection from point clouds},
  author={Lang, Alex H and Vora, Sourabh and Caesar, Holger and Zhou, Lubing and Yang, Jiong and Beijbom, Oscar},
  booktitle={Proceedings of the IEEE/CVF conference on computer vision and pattern recognition},
  pages={12697--12705},
  year={2019}
}

@inproceedings{yin2021center,
  title={Center-based 3d object detection and tracking},
  author={Yin, Tianwei and Zhou, Xingyi and Krahenbuhl, Philipp},
  booktitle={Proceedings of the IEEE/CVF conference on computer vision and pattern recognition},
  pages={11784--11793},
  year={2021}
}

@misc{mmdet3d2020,
    title={{MMDetection3D: OpenMMLab} next-generation platform for general {3D} object detection},
    author={MMDetection3D Contributors},
    howpublished = {\url{https://github.com/open-mmlab/mmdetection3d}},
    year={2020}
}

@inproceedings{dong2023lidar,
  title={Lidar-based cooperative relative localization},
  author={Dong, Jiqian and Chen, Qi and Qu, Deyuan and Lu, Hongsheng and Ganlath, Akila and Yang, Qing and Chen, Sikai and Labi, Samuel},
  booktitle={2023 IEEE Intelligent Vehicles Symposium (IV)},
  pages={1--8},
  year={2023},
  organization={IEEE}
}

@inproceedings{biber2003normal,
  title={The normal distributions transform: A new approach to laser scan matching},
  author={Biber, Peter and Stra{\ss}er, Wolfgang},
  booktitle={Proceedings 2003 IEEE/RSJ International Conference on Intelligent Robots and Systems (IROS 2003)(Cat. No. 03CH37453)},
  volume={3},
  pages={2743--2748},
  year={2003},
  organization={IEEE}
}

@inproceedings{besl1992method,
  title={Method for registration of 3-D shapes},
  author={Besl, Paul J and McKay, Neil D},
  booktitle={Sensor fusion IV: control paradigms and data structures},
  volume={1611},
  pages={586--606},
  year={1992},
  organization={Spie}
}

@inproceedings{sultan2003assessing,
  title={Assessing the safety benefit of automatic collision avoidance systems (during emergency braking situations)},
  author={Sultan, Beshr and McDonald, Mike},
  booktitle={Proceedings of the 18th International Technical Conference on the Enhanced Safety of Vehicle.(DOT HS 809 543)},
  year={2003}
}

@article{westhofen2021criticality,
  title={Criticality metrics for automated driving: A review and suitability analysis of the state of the art},
  author={Westhofen, Lukas and Neurohr, Christian and Koopmann, Tjark and Butz, Martin and Sch{\"u}tt, Barbara and Utesch, Fabian and Kramer, Birte and Gutenkunst, Christian and B{\"o}de, Eckard},
  journal={arXiv preprint arXiv:2108.02403},
  year={2021}
}

@article{shetty2021safety,
  title={Safety challenges for autonomous vehicles in the absence of connectivity},
  author={Shetty, Akhil and Yu, Mengqiao and Kurzhanskiy, Alex and Grembek, Offer and Tavafoghi, Hamidreza and Varaiya, Pravin},
  journal={Transportation research part C: emerging technologies},
  volume={128},
  pages={103133},
  year={2021},
  publisher={Elsevier}
}

@inproceedings{wang2023does,
  title={Does physical adversarial example really matter to autonomous driving? towards system-level effect of adversarial object evasion attack},
  author={Wang, Ningfei and Luo, Yunpeng and Sato, Takami and Xu, Kaidi and Chen, Qi Alfred},
  booktitle={Proceedings of the IEEE/CVF international conference on computer vision},
  pages={4412--4423},
  year={2023}
}

\end{document}